\pdfoutput=1

\documentclass[11pt]{article}

\usepackage{EMNLP2022}

\usepackage{times}
\usepackage{latexsym}
\usepackage[T1]{fontenc}

\usepackage[utf8]{inputenc}

\usepackage{microtype}

\usepackage{inconsolata}

\usepackage{hyperref}

\usepackage{tabularx}
\usepackage{float}
\usepackage{graphicx} 

\usepackage{listings}
\usepackage{tabularx}
\usepackage{adjustbox}

\usepackage{algorithm}
\usepackage{algpseudocode}
\usepackage{multirow}
\usepackage{enumitem}

\usepackage[normalem]{ulem}

\usepackage{amsmath}
\usepackage{amssymb}

\usepackage{comment}
\usepackage{booktabs}
    \newcolumntype{L}{>{\raggedright\arraybackslash}X}
\usepackage{xcolor}

\definecolor{mbc}{RGB}{255, 40, 40}
\definecolor{mda}{RGB}{0, 200, 0}
\definecolor{abc}{RGB}{0, 153, 51}
\definecolor{emc}{RGB}{0, 40, 255}

\title{Handling Ontology Gaps in Semantic Parsing}

\author{Andrea Bacciu\textsuperscript{1,2},
        Marco Damonte\textsuperscript{1},
        Marco Basaldella\textsuperscript{1},
        Emilio Monti\textsuperscript{1} \\
        \textsuperscript{1}Amazon\\
        \textsuperscript{2}Sapienza University of Rome\\
        \texttt{\{andbac, dammarco, mbbasald, monti\}@amazon.com}}

\begin{document}
\maketitle

\begin{abstract}
The majority of Neural Semantic Parsing (NSP) models are developed with the assumption that there are no concepts outside the ones such models can represent with their target symbols (closed-world assumption).
This assumption leads to generate hallucinated outputs rather than admitting their lack of knowledge.
Hallucinations can lead to wrong or potentially offensive responses to users.
Hence, a mechanism to prevent this behavior is crucial to build trusted NSP-based Question Answering agents.
To that end, we propose the Hallucination Simulation Framework (HSF), a general setting for stimulating and analyzing NSP model hallucinations. The framework can be applied to any NSP task with a closed-ontology.
Using the proposed framework and KQA Pro as the benchmark dataset, we assess state-of-the-art techniques for hallucination detection.
We then present a novel hallucination detection strategy that exploits the computational graph of the NSP model to detect the NSP hallucinations in the presence of ontology gaps, out-of-domain utterances, and to recognize NSP errors, improving the F1-Score respectively by $\sim$21\%, $\sim$24\% and $\sim$1\%.
This is the first work in closed-ontology NSP that addresses the problem of recognizing ontology gaps.\\
We release our code and checkpoints at {\small\url{https://github.com/amazon-science/handling-ontology-gaps-in-semantic-parsing}}.

\end{abstract}

\section{Introduction}\label{sec:introduction}

Semantic Parsing (SP) is one of the long-standing tasks in Natural Language Understanding, aiming at mapping complex natural language to machine-readable languages (e.g., SQL, SPARQL, KoPL \citep{cao2022kqa}, and so on). 
These languages, which we will refer to as Meaning Representation Languages (MRLs), are designed to be precise representations of the natural language's intent, enabling efficient querying of a Knowledge Base (KB) to retrieve pertinent answers in a Question Answering (QA) agent.
Despite the advent of the Transformer architecture \citep{vaswani2017attention}, which has enabled semantic parsers to achieve extraordinary performance \citep{cao2022kqa, bai2022graph, conia-etal-2021-unifying}, Semantic Parsing's crux remains the handling of out-of-ontology queries; in other words, since SP models and tasks (such as KQA-PRO~\citep{cao2022kqa}, LC-QUAD 2.0~\citep{dubey2019lc}, and QALD-9 \citep{cui2022compositional}) hold a closed-world assumption, they will always try to map an utterance to a MRL \emph{even if there is no valid representation for that utterance in the target ontology}, leading to wrong answers to be delivered to the model's users, called hallucinations. 

In fact, the closed-ontology task formulation enforces NSP models to always produce interpretations without an option to admit their lack of knowledge, inducing the models to hallucinate.
Therefore, the resulting models produce hallucinated outputs when they receive an utterance that requires symbols outside of their ontology, resulting in a wrong and potentially offensive answer. 
It is then of paramount importance to develop a system able to detect and prevent these hallucinations, so that users are not exposed to such mistakes.
Hallucinations in NSP differs with the notion of hallucinations in Natural Language Generation, we report the differences in Appendix \ref{app:nlg-vs-nsp}.
\begin{figure*}[h!]
\centering
\includegraphics[scale=0.40]{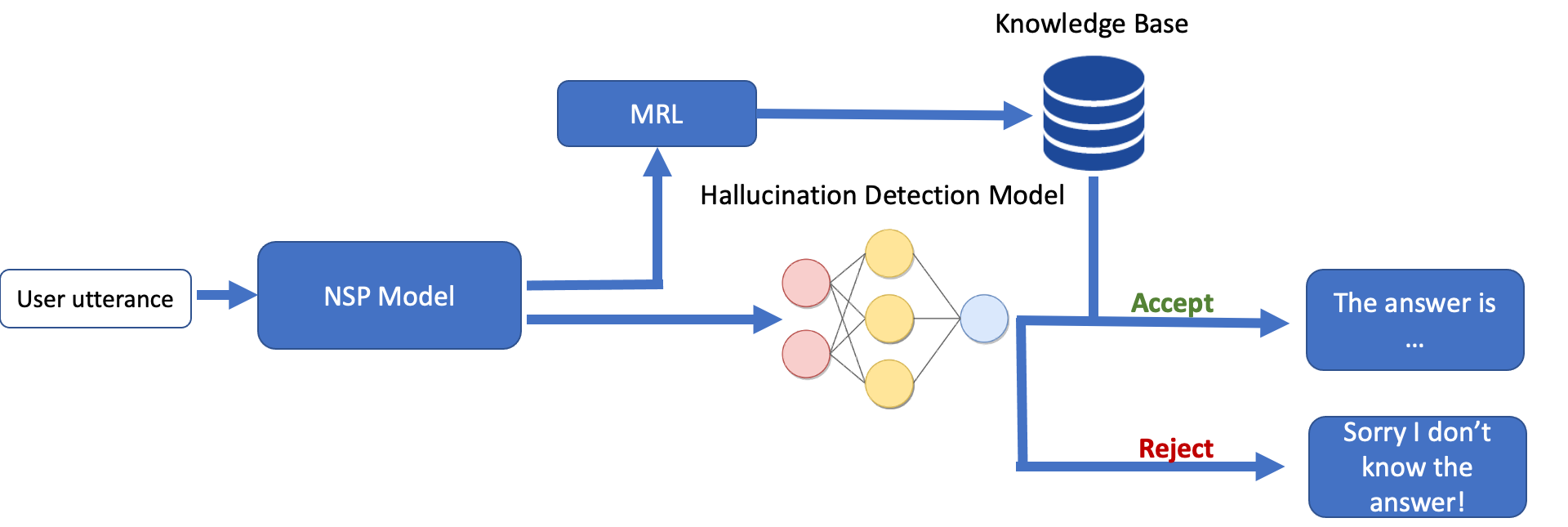}
\caption{The proposed pipeline: (1) the NSP model (KQA-PRO Bart model) receives the question from the user and produces the corresponding MRL; (2) the Hallucination Detection Model extracts features from the NSP model and decides whether to deliver the MRL to the user or not. }
\label{fig:e2e-pipeline}
\end{figure*}

To better understand different types of hallucinations in NSP, we classify errors into four macro categories. Given a semantic Q\&A parsing task $\mathcal{T}$, a dataset $\mathcal{D}$,
and an ontology $\mathcal{O}$, hallucinations of a model trained over $\mathcal{D}$ are classified as:
\begin{itemize}[noitemsep,topsep=0pt,parsep=0pt,partopsep=0pt]
    \item \textbf{in-ontology NSP errors}: utterances within the scope of $\mathcal{T}$ and where $\mathcal{O}$ contains all the symbols required to produce the correct MRLs, but for which the NSP model produces an incorrect MRL. For example, the utterance \textit{``What is the capital of France?"} is in-ontology if $\mathcal{O}$ contains the symbols for ``\textit{France}'' and ``\textit{capital of}''. However, if the NSP model erroneously translates the utterance to an MRL referencing e.g. a symbol for \emph{``weather of''} instead of \emph{``capital of''}, this type of hallucination is categorized as in-ontology NSP error. We will refer to this kind of errors as \textbf{NSP errors} for brevity.
    
    \item \textbf{out-of-ontology}: utterances that are within $\mathcal{T}$ but for which $\mathcal{O}$ does not contain all the symbols required to produce the correct MRLs (ontology gap). For example, \textit{``What is the crime rate of France?"}, is out-of-ontology if $\mathcal{O}$ does not contain a symbol for the predicate ``\textit{crime-rate-of}''. In this case, the NSP model will hallucinate another symbol, e.g. it could generate the MRL for \emph{``what is the population of France''} instead.
    
    \item \textbf{out-of-domain (OOD)}: utterances outside the scope of $\mathcal{T}$. For example, if $\mathcal{T}$ = factual QA, \textit{``Switch on the lights!''} is OOD because it is not a factual question. We expect an empty MRL because $\mathcal{O}$ does not have the necessary symbols to satisfy the out-of-ontology user utterance and the NSP model is trained to perform the task $\mathcal{T}$.

    \item \textbf{non-executable output}: in this case, the NSP model will output a MRL that cannot be executed and it thus cannot lead to an answer.
\end{itemize}
We show actual closed ontology Semantic Parsing hallucination examples in Figure \ref{fig:hallucinated_mrl} and we report more in Appendix \ref{app:second-hallucination}.
High performance in detecting OOD utterances in NSP can be achieved \citep{lukovnikov2021detecting, lang2023survey}, and non-executable outputs are trivially detectable as they fail to parse; but identifying both in-ontology and out-of-ontology errors can be hard even for experienced annotators, since the sheer size of most popular ontologies makes it impractical for a human to have a complete view of all the ontology symbols\footnote{e.g., Wikidata has 10k+ properties.}. Moreover, to the best of our knowledge, there are no works addressing this specific NSP issue. 

The research question that we want to address is: \emph{what is the most effective strategy to prevent a NSP-based QA agent to deliver wrong, and potentially offensive, answers to its users?} 
To this end, we develop the Hallucination Simulation Framework depicted in Figure \ref{fig:e2e-pipeline}; in detail, our main contributions are:
\begin{itemize}[noitemsep,topsep=0pt,parsep=0pt,partopsep=0pt]
    \item We propose a framework to stimulate, analyze and detect hallucinations in closed ontology NSP;
    \item We propose the Hallucination Detection Model (HDM), an architecture that analyzes an NSP model to determine whether it is hallucinating or not using several hallucination detection signals;
    \item We introduce a model's \emph{Activations} as hallucination detection signals; when combined with other signals, they improve the Macro F1-Score by up to 21\% in ontology gaps, 1\% in NSP error, and 24\% in OOD detection.
\end{itemize}
To the best of our knowledge, this is the first work that addresses the ontology gaps problem in a closed-ontology NSP task.

\section{Related Work}\label{sec:related-work}
When we do not allow models to admit their lack of knowledge, forcing them to produce an output even when they do not have the instruments to do it, they will inevitably \emph{hallucinate}.
In other words, in generative NLP, when the generated output displays a misunderstanding of the input utterance by the model, we say that the model is ``hallucinating''.
Typically, models hallucinate in two ways: (1) inventing additional information not included in or related to the input utterance, or (2) confusing a symbol/word with another one.

One of the biggest assumptions in existing Semantic Parsing tasks is that every input always has a valid target logical form. 
In such a setup, models are always forced to generate a MRL or, in other words, to hallucinate a wrong understanding, instead of admitting a lack of knowledge.
However, recently the NLP community has begun to investigate this closed-world assumption for other tasks.
For example, the Extractive Question Answering dataset SQuAD v1 \citep{rajpurkar2016squad} was built with the assumption that, given each question-paragraph pair, it is always possible to find an answer to the question in the paragraph.
This assumption was removed in the second version of the dataset \citep{rajpurkar2018know}, which includes questions without an answer.
Another field in which this problem was addressed is entity linking, where models can produce a \textsc{nil} entity when they cannot find a suitable entity for a certain mention \citep{ruas2022nilinker}.
On the other hand, most of the hallucination detection in NSP works rely on two confidence estimation techniques: (1) the Sequence log-probability (also called Confidence Score) \citep{guerreiro2022looking, dong2018confidence}, or (2) Monte Carlo Dropout (or Dropout Perturbation) \citep{gal2016dropout,guerreiro2022looking, dong2018confidence}.

\section{Closed World Assumption in NSP: A Logical Theory Perspective} \label{sec:cwa}
The Closed World Assumption (CWA) originates from logic theory, and it is the assumption that only the known facts are correct, and what is not known is false \citep{reiter1981closed, keet2013closed}. In other words, the CWA assumes \emph{total knowledge} over a domain, implying that all the possible symbols (e.g., entities and predicates) are known, and that only the known facts represented using the known symbols are true. On the other hand, the Open World Assumption (OWA) makes no assumption over what is not known; in other words, the OWA allows ``gaps'' in the knowledge, e.g. the existence of unknown symbols or of unknown, but true, facts.

For some tasks, using the CWA is safe. For example, \citet{reiter1981closed} notes that: ``in an airline data base, all flights and the cities which they connect will be explicitly represented. If I fail to find an entry indicating that Air Canada flight 103 connects Vancouver with Toulouse I will conclude that it does not''. For SP models, however, the CWA can be dangerous. Let's take the following scenario: a CWA NSP model's input is \emph{``what is the crime rate of France''}, but the target ontology does not have a representation for the predicate \emph{``crime rate of''}. Since the model is trained under the CWA it assumes that there cannot be other predicates other the ones it can access, hence it will (a) ground \emph{``crime rate of''} to a different predicate and then (a) produce a necessarily incorrect representation of the input. If this incorrect representation happens to be syntactically correct, it will then be exectuted, serving a wrong answer to the customer.





\begin{figure}
\centering
\includegraphics[width=0.49\textwidth]{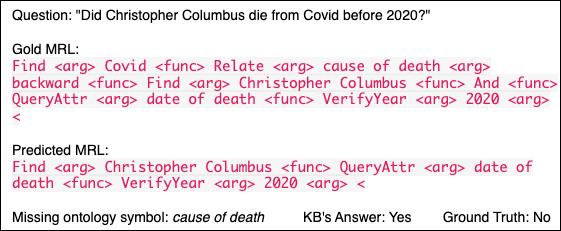}
\caption{We show the output our NSP model trained without a symbol for the concept of ``cause of death". Given a 
question that requires this symbol, the model produces a wrong but executable MRL 
leading to a wrong answer served to its user. 
}
\label{fig:hallucinated_mrl}
\end{figure}

This issue is exemplified in Figure \ref{fig:hallucinated_mrl}, where we take a NSP model trained on the KQA-Pro dataset \citep{cao2022kqa} and we ask it to generate a MRL for the question \emph{``did Chistopher Columbus die from Covid before 2020?''}. The absence of the "cause of death" symbol in the set of the model's known symbols leads to an MRL which erroneously uses the "date of death" symbol instead. Even if this MRL is syntactically correct, it misrepresents the input due to the limitations of the training set. Since the MRL is executable, it will lead to the generation of an incorrect answer.

\section{Detecting NSP Hallucinations}\label{sec:hallucination-simulation-framework}

\subsection{Hallucination Simulation Framework}
\label{sec:hsf}
Building on the CWA and OWA assumptions, we introduce the Hallucination Simulation Framework (HSF), a dataset-agnostic approach tailored for closed-ontology NSP tasks. 
This framework leverages the closed and open world assumptions to force a model to hallucinate at inference time. The model is trained using a ``normal'' SP dataset holding the CWA. However, the validation and test sets will contain MRLs needing symbols not known to the model at training time, hence forcing it to hallucinate. This allows to analyse how the model behaves when unable to produce ontology symbols, and to develop a number of hallucination detection strategies to mitigate the issue.

In practical terms, the HSF operates by considering the ontology used for a CWA SP dataset 
\(\mathcal{O}_{\text{dataset}}\), and decomposing it into two disjoint sub-ontologies, called 
\(\mathcal{O}_{\text{known\_symbols}}\) and \(\mathcal{O}_{\text{unknown\_symbols}}\). 
\(\mathcal{O}_{\text{known\_symbols}}\) contains the ontology symbols that are used to train the model, while \(\mathcal{O}_{\text{unknown\_symbols}}\) contains the symbols that are used to stimulate hallucinations; we have that \(\mathcal{O}_{\text{known\_symbols}} \cup \mathcal{O}_{\text{unknown\_symbols}} = \mathcal{O}_{\text{dataset}}\) and \(\mathcal{O}_{\text{known\_symbols}} \cap \mathcal{O}_{\text{unknown\_symbols}} = \varnothing \).

These sub-ontologies are used to construct two datasets, a \textit{NSP dataset} and an \textit{Hallucination Detection Dataset} (HDD), whose construction is detailed in Section \ref{sec:hdd}. The NSP dataset, containing only \(\mathcal{O}_{\text{known\_symbols}}\), is used to train the model, while the HDD, containing both \(\mathcal{O}_{\text{known\_symbols}}\) and 
\(\mathcal{O}_{\text{unknown\_symbols}}\), is used to stimulate the model to hallucinate wrong ontology symbols and develop hallucination detection strategies (Section \ref{sec:hallucinations-features}) . 

Thanks to this framework, we can now programmatically induce hallucinations in a NSP model at inference time. Thus, we can train, tune, and test hallucination detection strategies to recognize unwanted signals from the model. 

\label{sec:hdd}
\subsection{Hallucination Detection Dataset}
The HDD comprises two types of samples: each natural language sentence is paired either with (1) MRLs that require only symbols from the $\mathcal{O}_{\text{known\_symbols}}$  set, or (2) MRLs that require at least one symbol from the $\mathcal{O}_{\text{unknown\_symbols}}$ set.
To build the HDD, we first define $\mathcal{O}_{\text{unknown\_symbols}}$;
then, we split $\mathcal{O}_{\text{unknown\_symbols}}$ in three sets, that are used to build the HDD train, development, and test set.
We report the complete $\mathcal{O}_{\text{unknown\_symbols}}$ set in Appendix \ref{app:out-of-ontology-symbol-list}. 

In the following, we describe the methodology we followed to we ensure that the out-of-ontology symbols are sufficiently diverse and challenging, providing a rigorous test of the hallucination detection strategies.

\paragraph{Disjoint HDD train, validation and test sets}\label{sec:out-of-ontology-split}
To ensure that $\mathcal{O}_{\text{unknown\_symbols}}$ cannot be shared across train, dev and test set, we create three disjoint set one for each data split, as shown in Appendix \ref{app:out-of-ontology-symbol-list}.
Furthermore, we eliminate any sentences that require symbols from multiple out-of-ontology splits. This allows the development of robust hallucination detection strategies that are able to generalise over unseen ontology symbols.

\paragraph{Diversification of unknown symbols}\label{sec:tail-ontology}
To improve the generalization of our methods, we also aim to maximize the number of out-of-ontology symbols across all splits. 
This is essential, as having few unknown symbols might lead hallucination detection strategies to recognize them throw their sentence context than isolating the underlying hallucination signal.
For this purpose, we place symbols in $\mathcal{O}_{\text{unknown\_symbols}}$ based on their frequency of occurrence within the original dataset; we prioritize symbols with lower frequency (symbols with maximum 2 occurrences), as this approach maximizes the number of unique symbols in the HDD while maintaining a robust volume of samples for the NSP training set.

\paragraph{Ensuring Independent Feature Extraction by Dataset Segregation}\label{sec:in-ontology-bias}
As detailed above, the framework employs two datasets: the NSP dataset and the HDD, each divided into training, dev, and test splits.

To construct the known symbol portion of the HDD we used utterances from the NSP dataset.
It is crucial not to include utterances from the NSP train split, otherwise the hallucination detection strategies could simply learn to recognize as non-hallucinated only the utterances that were used to train the NSP model.

\begin{figure*}
\centering
\includegraphics[scale=0.40]{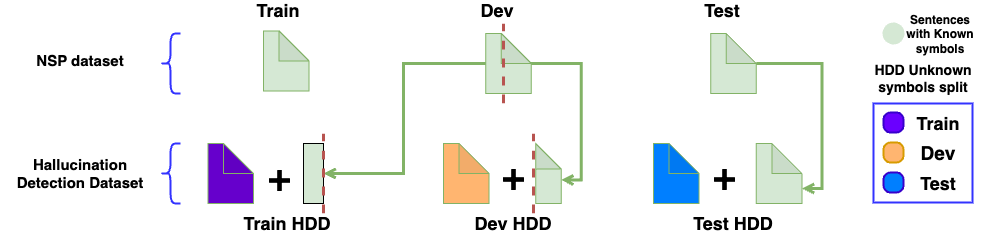}
\caption{Construction of the Hallucination Detection Dataset (HDD).
The first row represents the dataset used to train and test the NSP model, containing only $\mathcal{O}_{\text{known\_symbols}}$.
To construct the $\mathcal{O}_{\text{known\_symbols}}$ portion of the HDD while avoiding overfitting of the hallucination detection strategies, we sourced sentences only from the validation and test splits of the NSP dataset as explained in Section \ref{sec:in-ontology-bias}.}
\label{fig:hdd_construction}
\end{figure*}

To circumvent this issue, the training and validation sets of the HDD are built by splitting NSP validation set.
The HDD test set is instead simply built by appending samples containing the test symbols from $\mathcal{O}_{\text{unknown\_symbols}}$ to the existing NSP test set.
We depict this process in Figure \ref{fig:hdd_construction}.

\paragraph{Out-Of-Domain sentences} 
Besides out-of-ontology sentences, also out-of-domain (OOD) sentences are a common cause of hallucinations for NSP models.
For example, consider a system trained to answer questions like ``In what state does the Pope live?". Given an input sentence such as ``Set an alarm at 8 am for Monday!" from a distinct domain (i.e., not a question), the question answering system will always produce wrong MRLs, because its ontology is not suitable for this type of utterances.
We include OOD only in the validation and test sets for two main reasons: 1) to evaluate the zero-shot capabilities in recognizing OOD utterances as a different source of out-of-ontology; 2) to avoid the need for specific training for OOD detection, as addressing the wide range of potential OOD instances is beyond the scope of this study.
We report the OOD dataset statistics in Appendix \ref{app:ood-stats}.

\section{Hallucination Detection Strategies}\label{sec:hallucinations-features}
In this Section, we introduce the Hallucination Detection Strategies that we use in our experiments.
\paragraph{Autodetect Hallucinations}
A baseline approach to detect hallucinations is to enable the NSP model itself to decide whether to reject the MRL or not, in a similar fashion to the NIL entity in \citet{ruas2022nilinker}.
Therefore, we add a new ontology symbol called \texttt{<Reject-MRL>} in the NSP model, as a label for all the out-of-ontology sentences, i.e. moving from a CWA approach to a OWA one. 
Instead of using the NSP and HDD datasets, as we don't rely on external hallucination detection strategies, we train the NSP model using the full  $\mathcal{O}_{\text{dataset}}$, marking MRLs containing  $\mathcal{O}_{\text{unknown\_symbols}}$ samples as utterances to reject.
In preliminary experiments, this approach resulted in zero true positives.
This happens because the model memorized the utterances marked as out-of-ontology, hence failing to generalize on the ``unseen'' unknown symbols in the development and test set (see Section \ref{sec:out-of-ontology-split} for how the disjoint train, validation and test sets are constructed).



\begin{figure}[h]
    \includegraphics[scale=0.49]{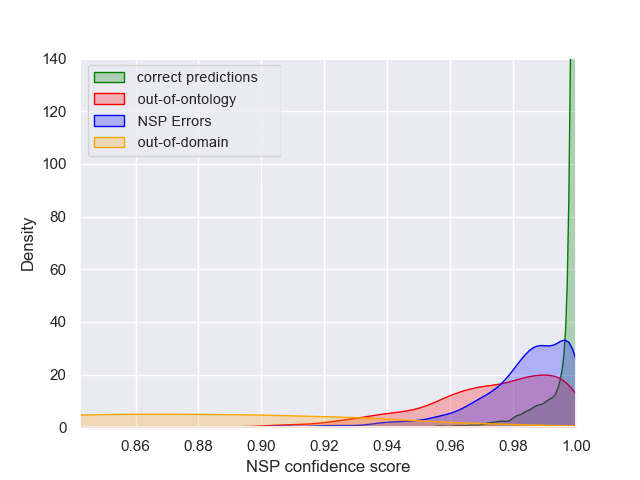}
    \caption{Overlap between the distributions of correct predictions, out-of-ontology, NSP errors, and OOD w.r.t. Confidence Score (CS). The model is overconfident over wrong predictions, hence the CS is not sufficient to separate good and Hallucinated MRLs. Specifically, the CS struggles to distinguish between NSP Errors and correct predictions (i.e., both types of MRLs that contains only $\mathcal{O}_{\text{known\_symbols}}$).}
    \label{fig:perplexity-distribution}
\end{figure}

\paragraph{Confidence Score}
Confidence Score (CS) is a standard method to detect hallucinations \citep{dong2018confidence} that measures the confidence level of a statistical model about the output it generates.
However, this method relies on the strong assumption that the model will not be confident when generating hallucinations, and vice versa. This is not always guaranteed in practice: as we can see in the CS distribution in Figure \ref{fig:perplexity-distribution}, the confidence distributions of correct and wrong model predictions overlap. For this reason, rejecting model predictions below a certain threshold would not be sufficient to remove all the wrong MRLs.

To compute the CS, we calculate the Posterior Probability (PP) of a generated MRL $w_{n}, ..., w_{1}$ from the beam search tree, and then we normalize it by the length $n$ of the generated output, by applying the $n$th-root.
\begin{equation}
    CS = \sqrt[n]{PP(w_{n}, w_{n-1}, ..., w_{1})}
\end{equation}
We test CS in two ways: (1) by setting a threshold to the best CS value found in a sample from the HDD train set that maximizes the hallucination detection in the HDD dev set; (2) and by using it as a feature in the Hallucination Detection Model (HDM) that we will define in Section \ref{sec:experimental-setup}.

\paragraph{Monte Carlo Dropout}
The Monte Carlo Dropout (MCD) strategy was introduced by \citet{gal2016dropout}: the idea is to use the dropout technique as a Bayesian approximation to represent the model uncertainty.
Dropout is a well-known regularization technique that randomly disables a subset of the neurons in a neural network layer in order to prevent overfitting.
MCD involves enabling dropout at inference time and running inference multiple times to create a random perturbation in the model; a small perturbation indicates that the model is confident with the input, while a large perturbation suggests a likely mistake from the model.
We follow the formulation by \citet{dong2018confidence}, using 30 trials, beam size of 2, and taking the variance of the CS value.
Similar to CS, we use the MCD in two ways: (1) identifying a threshold value that maximizes hallucination detection between out-of-ontology/NSP Errors and in-ontology, and (2) using it as a feature for the HDM.

\paragraph{Model Activations}
Looking at the activations of the model's computational graph is a powerful way to debug neural networks and is usually used for explainability, such as in the Grad-Cam algorithm \citep{selvaraju2017grad}.
For this reason, we propose for the first time to use the forward activations of the NSP model encoder at inference time to detect whether there is a hallucination or not.
To encode the activation features for all layers, we pool the sequence length and compute the variance.
Then, we use the encoding of the model's activations as a feature to recognize the hallucinations in the HDM. 
Although it can be argued that both the Autodetect and Activations strategies use the encoder's hidden states, these approaches are different. 
The first approach uses only the last hidden states of the encoder as input to the decoder, which has then the duty of producing an MRL or the rejection symbol.
On the other hand, in the HDM all the encoder's activations are used as input, allowing the HDM to have a complete view of the hidden states of the NSP model during the generation. 

\paragraph{Hallucination Detection Model}\label{sec:hallucination-detection-model-exp}
The Hallucination Detection Model (HDM) is a neural network trained on the HDD that learns to classify whether an NSP model is hallucinating or not using as features the signals extracted from the NSP models, such as CS, activations, and MCD. 
The HDM consists of a MultiHead-Attention and two feed-forward layers with RELU function, batch normalization, dropout, and a binary classification head.
We report a Figure of the architecture in Appendix \ref{hdm-architecture}, the complete list of hyper-parameters in Appendix \ref{app:hdm-config} and hardware infrastructure in Appendix \ref{app:hardware}.



\section{Experimental Setup}\label{sec:experimental-setup}
\paragraph{Dataset} 
While the HSF is dataset-agnostic, in our experiments, we use the KQA-PRO dataset \citep{cao2022kqa}, based on the KoPL (Knowledge-oriented Programming Language) MRL; this dataset is built on top of a large ontology, which is a subset of Wikidata. We instead sourced OOD sentences from the TOP v2 Dataset \citep{chen2020low}, that contains task oriented utterances, such as ``Turn on the lights!". 

To create a test set, we merged the train and the validation set, and we split the data as follows: 60\%, 20\%, and 20\%, respectively, for the train, validation and test set.
The statistics of the HSF framework applied to the KQA-PRO dataset are reported in Appendix \ref{app:hsf-dataset-stats}.

\paragraph{NSP model} 
Following the KQA-PRO paper, we train the BART-base model \citep{lewis2019bart}, using the NSP training dataset. We report the hyper-parameters that we use to train the NSP model in Appendix \ref{app:kqa-pro-hparams}. Note that as the original KQA-PRO test set is not publicly available, we cannot compare our results with the original dataset paper.

\paragraph{Evaluation}
To measure the hallucination detection capabilities, we use the Macro F1-Score due to the imbalance of the dataset, as shown in Appendix \ref{app:ood-stats} and, \ref{app:out-of-ontology-symbol-list}.
We compute the individual F1-Score for each type of hallucination defined in Section \ref{sec:introduction}: in-ontology \textbf{NSP errors} caused by the model hallucinating wrong symbols from $\mathcal{O}_{\text{known\_symbols}}$, \textbf{out-of-ontology errors} caused by the need of symbols in $\mathcal{O}_{\text{unknown\_symbols}}$ to correctly represent the input, and zero-shot \textbf{OOD} detection. 
As mentioned in Section \ref{sec:executable-non-executable}, we excluded non-Executable MRLs from our evaluation protocol because they are trivially detected by simply trying, and failing, to execute them on the KB. To increase the robustness of our results, we repeat the training of the HDM model in all the configurations using 10 different random seeds, and then we report the mean and the standard deviation of the F1-Scores.

\begin{table}
\begin{center}
\resizebox{\columnwidth}{!}{%
\begin{tabular}[width=0.50\textwidth]{lcc}
    \toprule
    \textbf{Split} & \textbf{Answer Accuracy} & \textbf{MRL EM} \\ 
    \midrule
    NSP model (baseline) & 93\% & 85\%  \\ 
    NSP model + Threshold CS  & 96\% & 94\% \\
    NSP model + Act. + CS & \textbf{97\%} & \textbf{95\%} \\
    \bottomrule
\end{tabular}
}%
\end{center}
\caption{
Performance of baseline KQA-PRO BART model and of the best hallucination detection models on the NSP task; the NSP model is trained as in \citep{cao2022kqa}, and on top of it we apply our hallucination detection strategies. We compute metrics only on the executable outputs that lead to an answer to be delivered to a user; for more details, see Appendix \ref{app:metrics}.
}\label{tab:KQA-PRO-results} 
\end{table}



\begin{figure*}
    \centering
    \includegraphics[scale=0.36]{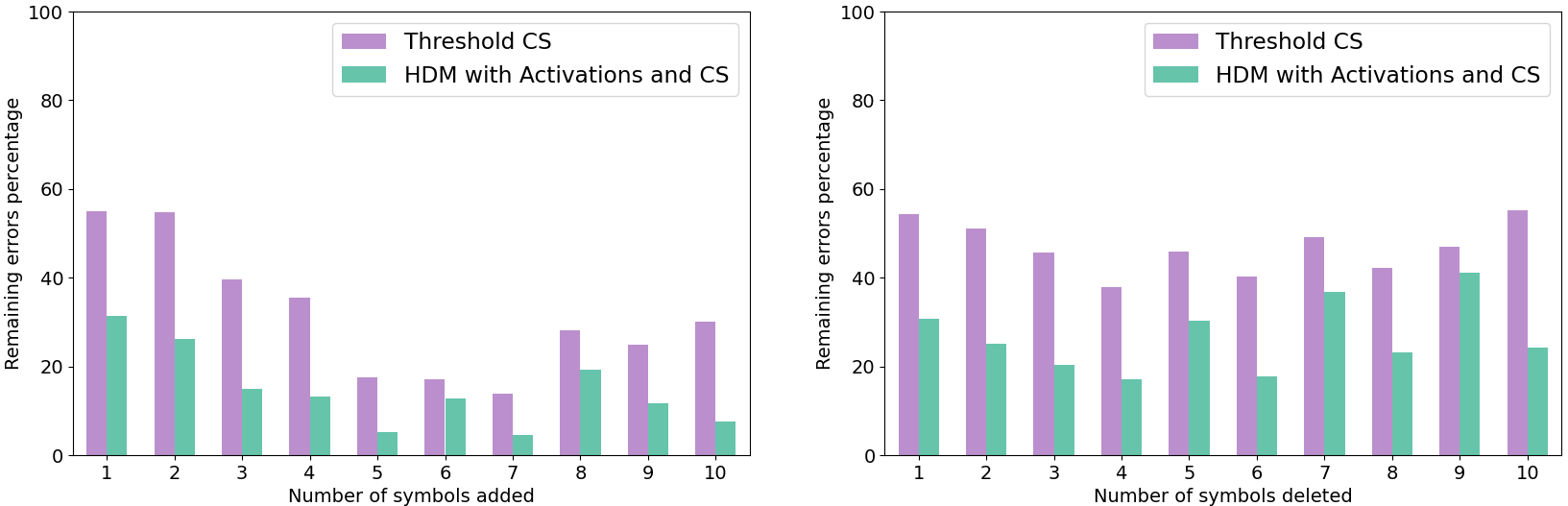}
    \caption{In this plot on the y-axis the percentage of remaining error ($\downarrow$ is better) and on the x-axis we distinguish between the various hallucinated MRLs that omit (right plot) or add (left plot) incorrect ontology symbols with respect to the ground truth. Residual error compares two systems: Threshold CS and HDM with Activations and CS.}
    \label{fig:error-analysis}
\end{figure*}

\section{Discussion}\label{sec:discussion}
We report the performance of our NSP model using Execution Accuracy and the MRL Exact Match metric in Table~\ref{tab:KQA-PRO-results}.
In this work, we focused on four major causes for hallucinations: in-ontology NSP errors, out-of-ontology utterances and out-of-domain utterances.
Specifically, we propose the first work that addresses the problem of ontology gaps, i.e., exposing an NSP model to utterances that require unknown ontology symbols to be represented in the output vocabulary. As mentioned in Section \ref{sec:introduction}, recognizing ontology gaps is a challenging task even for experienced annotators due to the large size of the most popular ontologies.
Our methodology induces ontology gaps and forces the model to hallucinate programmatically through a Hallucination Simulation Framework (\S\ref{sec:hallucination-simulation-framework}). We developed a number of hallucination prevention strategies (\S\ref{sec:hallucinations-features}) to detect and prevent the delivery of hallucinated answers to users.
In Table \ref{tab:performance}, we report the individual Macro F1-Score of the tested systems on the three scenarios: out-of-ontology, NSP Errors, and zero-shot out-of-domain.

From a baseline where only non-executable MRLs are not delivered to the user, the HDM with \textit{Activations + CS} is our best-performing model, improving answer accuracy by 4\% and MRL exact match by 10\%, effectively reducing a user's exposition to wrong answers.
The HDM with \textit{Activations + CS}'s performance is achieved by 
increasing the Macro F1-Score by approximately 21\% and 24\% for out-of-ontology and out-of-domain detection w.r.t. baseline, respectively. On the other hand, the NSP Errors detection performance is comparable to that of Threshold CS, with only the HDM with the \textit{Activations + CS + MCD} combination showing a 1\% improvement over the baseline in NSP Error detection. This marginal gain can be attributed to the limited number of errors produced by our NSP model over utterances with known symbols only, which constitutes about 11\% of the in-ontology utterances (see statistics in Appendix \ref{app:analysis-nsp-errors}).

However, we can notice that both CS and MCD, if optimized through the HDM, obtain large gains in terms of Macro F1-Score. In fact, CS improves by 17\% and 10\% in out-of-ontology and out-of-domain detection, and MCD by 4\%, 3\%, and 10\% in out-of-ontology, NSP Errors, and out-of-domain detection.
In addition, the HDM can combine multiple hallucination signals to obtain higher performance, as in the case of our most-performing system.
For further insight, we report the Precision and Recall over each error category in Appendix \ref{app:precision-recall}.

\begin{table*}[t]
    \centering
    \resizebox{0.89\textwidth}{!}{%
    \begin{tabular}[t]{lcccc}
    \toprule
    \textsc{\textbf{Exp name - End2End}} & \textbf{out-of-ontology} & \textbf{NSP error} & \textbf{out-of-domain} & \textbf{average} \\
    \midrule
        Autodetect (Baseline) & 0.490 & 0.471 & 0.466 & 0.476 \\
        Threshold CS 98.5\% (Baseline) & 0.480 & 0.653 & 0.456 & 0.530 \\
        Threshold MCD (Baseline) & 0.452 & 0.439 & 0.428 & 0.440 \\
        Activations$^{HDM}$ & 0.498 $\pm$ 0.013 & 0.474 $\pm$ 0.003 & 0.466 $\pm$ 0.003 & 0.479 \\
        CS$^{HDM}$ & 0.648 $\pm$ 0.040 & 0.591 $\pm$ 0.023 & 0.552 $\pm$ 0.089 & 0.597 \\
        MCD$^{HDM}$ & 0.490 $\pm$ 0.001 & 0.471 $\pm$ 0.003 & 0.541 $\pm$ 0.163 & 0.501 \\
        CS + MCD$^{HDM}$ & 0.654 $\pm$ 0.021 & 0.617 $\pm$ 0.018 & 0.537 $\pm$ 0.030 & 0.603 \\
        Activations + CS$^{HDM}$ & \textbf{0.701 $\pm$ 0.030} & 0.643 $\pm$ 0.027 & \textbf{0.703 $\pm$ 0.086} &\textbf{ 0.682} \\
        Activations + MCD$^{HDM}$ & 0.496 $\pm$ 0.012 & 0.474 $\pm$ 0.004 & 0.466 $\pm$ 0.002 & 0.479 \\
        Activations + CS+ MCD$^{HDM}$ & 0.659 $\pm$ 0.026 & \textbf{0.660 $\pm$ 0.025} & 0.618 $\pm$ 0.077 & 0.646 \\
    \bottomrule
\end{tabular}
    }%
    \caption{We report the Macro F1-Score ($\uparrow$ is better) in the three scenarios: out-of-ontology detection, NSP Error detection and zero-shot OOD detection. 
    These features are combined (+) concatenating their vector representations. The superscript $^{HDM}$ indicates the system optimized with the HDM.}
    \label{tab:performance}
    
\end{table*}

\paragraph{Executable vs Non-Executable MRLs}\label{sec:executable-non-executable}
To highlight the scale of the issue we are tackling, it is important to measure how many times wrong answers would be served to users without a proper hallucination detection pipeline. As shown in Appendix \ref{app:executableMRL-stats}, in 46.3\% of the utterances requiring 
\(\mathcal{O}_{\text{unknown\_symbols}}\) the NSP model generates a syntactically valid MRL, which would then be executed, causing a wrong answer to be delivered to the user. This happens because NSP models tend to replicate executable patterns using known symbols from the training set, even when receiving utterances that cannot be represented with the known vocabulary.

\paragraph{Effect of the number of changed ontology symbols}\label{sec:error_analysis}
To further analyze the results, we analyze the behavior of the NSP model on the hallucinated MRLs in Figure \ref{fig:error-analysis}.
Specifically, this analysis highlights the MRLs where the NSP model added wrong ontology symbols (left plot), or omitted required symbols (right plot) from the ground truth sequence. 
In the Figure, we show a comparison between two systems: (1) Threshold CS (the best non-model based strategy) and (2) HDM with Activations and CS (our best model-based strategy), expressed as a percentage of errors in relation to the number of modified symbols.
The plots suggest that (a) when the model \emph{adds} symbols, the hardest errors to detect happen when the model adds up to 2 unnecessary symbols, leaving $\ge 50\%$ of the errors undetected for CS and $\ge 30\%$ for the HDM; (b) when the model \emph{removes} symbols, there seems to be no discernible pattern based on the amount of removed symbols; and (c) in both cases, the HDM model performs considerably better than the Threshold CS strategy, with a relative error reduction of $\sim$50\%.


\paragraph{Latency} 
While adding a second neural network in the QA pipeline might be considered penalising in terms of latency, it's worth noting that the HDM is very small model compared to the main NSP model.
In detail, the HDM requires only 184k Floating Point Operations (FLOPs), which amounts to less than 1\% of the FLOPs required by the \textit{BART-base} architecture of the NSP model, which is 2.49 Billion FLOPs.

\section{Conclusions}\label{sec:conclusions}
Current studies of Neural Semantic Parsing (NSP) models revolve around improving performance on academic benchmarks, but they do not take into account the trustworthiness of the model in a real world scenario where the model is used to serve answers to users of a QA system.
In such scenario, NSP models can hallucinate syntactically correct, but semantically wrong MRLs, that can be used to serve incorrect answers to users. This is particularly true when users ask questions that require knowledge beyond the one used by the model's target ontology, as in these cases the model simply cannot generate a correct MRL.

To test NSP models under this more realistic scenario, we propose the Hallucination Simulation Framework (HSF), where we programmatically induce NSP models to hallucinate, and then, using the Hallucination Detection Model, we detect model errors at inference time using several different signals, including the model's activations or Confidence Score, or by using Monte Carlo Dropout. 

We find that the best way to prevent detect hallucinations is using the HDM model with Activations and CS as features, which leads to an average improvement of more than 20\% w.r.t. a baseline where the only non-served MRLs are just the syntactically incorrect ones.

\

\section*{Limitations}
There are some limitations in this work that do not concern the framework construction.
First of all, the framework imposes the construction of two datasets leading to a strong reduction of the training data. Hence, the framework to work properly requires a larger dataset.
We are eager to expand our work in the future by taking advantage of the proposed framework in the following directions:
(1) We pooled the activation sequences and did not take full advantage of the information in the sequences.
(2) We have not tested the individual probability of each token in the generated MRL.
(3) We have not tested the HDM with a multi-class output differentiating between in-ontology, out-of-ontology, NSP Errors, and OOD. (4) We did not test with other datasets, ontologies, or MRLs. (5) Our work has not been tested with other seq2seq architectures (e.g., mT5, Bart-large) and provides no multilingual tests.

\section*{Acknowledgments}
We are thankful to our colleagues for their feedback and suggestions during the completion of this work. We also appreciate our anonymous reviewers for their time and valuable comments.


\bibliography{anthology,custom}

\begin{thebibliography}{19}
\expandafter\ifx\csname natexlab\endcsname\relax\def\natexlab#1{#1}\fi

\bibitem[{Bai et~al.(2022)Bai, Chen, and Zhang}]{bai2022graph}
Xuefeng Bai, Yulong Chen, and Yue Zhang. 2022.
\newblock Graph pre-training for amr parsing and generation.
\newblock \emph{arXiv preprint arXiv:2203.07836}.

\bibitem[{Cao et~al.(2022)Cao, Shi, Pan, Nie, Xiang, Hou, Li, He, and Zhang}]{cao2022kqa}
Shulin Cao, Jiaxin Shi, Liangming Pan, Lunyiu Nie, Yutong Xiang, Lei Hou, Juanzi Li, Bin He, and Hanwang Zhang. 2022.
\newblock Kqa pro: A dataset with explicit compositional programs for complex question answering over knowledge base.
\newblock In \emph{Proceedings of the 60th Annual Meeting of the Association for Computational Linguistics (Volume 1: Long Papers)}, pages 6101--6119.

\bibitem[{Chen et~al.(2020)Chen, Ghoshal, Mehdad, Zettlemoyer, and Gupta}]{chen2020low}
Xilun Chen, Asish Ghoshal, Yashar Mehdad, Luke Zettlemoyer, and Sonal Gupta. 2020.
\newblock Low-resource domain adaptation for compositional task-oriented semantic parsing.
\newblock \emph{arXiv preprint arXiv:2010.03546}.

\bibitem[{Conia et~al.(2021)Conia, Bacciu, and Navigli}]{conia-etal-2021-unifying}
Simone Conia, Andrea Bacciu, and Roberto Navigli. 2021.
\newblock \href {https://doi.org/10.18653/v1/2021.naacl-main.31} {Unifying cross-lingual semantic role labeling with heterogeneous linguistic resources}.
\newblock In \emph{Proceedings of the 2021 Conference of the North American Chapter of the Association for Computational Linguistics: Human Language Technologies}, pages 338--351, Online. Association for Computational Linguistics.

\bibitem[{Cui et~al.(2022)Cui, Aralikatte, Lent, and Hershcovich}]{cui2022compositional}
Ruixiang Cui, Rahul Aralikatte, Heather Lent, and Daniel Hershcovich. 2022.
\newblock Compositional generalization in multilingual semantic parsing over wikidata.
\newblock \emph{Transactions of the Association for Computational Linguistics}, 10:937--955.

\bibitem[{Dong et~al.(2018)Dong, Quirk, and Lapata}]{dong2018confidence}
Li~Dong, Chris Quirk, and Mirella Lapata. 2018.
\newblock Confidence modeling for neural semantic parsing.
\newblock \emph{arXiv preprint arXiv:1805.04604}.

\bibitem[{Dubey et~al.(2019)Dubey, Banerjee, Abdelkawi, and Lehmann}]{dubey2019lc}
Mohnish Dubey, Debayan Banerjee, Abdelrahman Abdelkawi, and Jens Lehmann. 2019.
\newblock Lc-quad 2.0: A large dataset for complex question answering over wikidata and dbpedia.
\newblock In \emph{International semantic web conference}, pages 69--78. Springer.

\bibitem[{Gal and Ghahramani(2016)}]{gal2016dropout}
Yarin Gal and Zoubin Ghahramani. 2016.
\newblock Dropout as a bayesian approximation: Representing model uncertainty in deep learning.
\newblock In \emph{international conference on machine learning}, pages 1050--1059. PMLR.

\bibitem[{Guerreiro et~al.(2022)Guerreiro, Voita, and Martins}]{guerreiro2022looking}
Nuno~M Guerreiro, Elena Voita, and Andr{\'e}~FT Martins. 2022.
\newblock Looking for a needle in a haystack: A comprehensive study of hallucinations in neural machine translation.
\newblock \emph{arXiv preprint arXiv:2208.05309}.

\bibitem[{Keet(2013)}]{keet2013closed}
C~Maria Keet. 2013.
\newblock Closed world assumption.
\newblock \emph{Encyclopedia of Systems Biology}, pages 415--415.

\bibitem[{Lang et~al.(2023)Lang, Zheng, Li, Sun, Huang, and Li}]{lang2023survey}
Hao Lang, Yinhe Zheng, Yixuan Li, Jian Sun, Fei Huang, and Yongbin Li. 2023.
\newblock A survey on out-of-distribution detection in nlp.
\newblock \emph{arXiv preprint arXiv:2305.03236}.

\bibitem[{Lewis et~al.(2019)Lewis, Liu, Goyal, Ghazvininejad, Mohamed, Levy, Stoyanov, and Zettlemoyer}]{lewis2019bart}
Mike Lewis, Yinhan Liu, Naman Goyal, Marjan Ghazvininejad, Abdelrahman Mohamed, Omer Levy, Ves Stoyanov, and Luke Zettlemoyer. 2019.
\newblock Bart: Denoising sequence-to-sequence pre-training for natural language generation, translation, and comprehension.
\newblock \emph{arXiv preprint arXiv:1910.13461}.

\bibitem[{Lukovnikov et~al.(2021)Lukovnikov, Daubener, and Fischer}]{lukovnikov2021detecting}
Denis Lukovnikov, Sina Daubener, and Asja Fischer. 2021.
\newblock Detecting compositionally out-of-distribution examples in semantic parsing.
\newblock In \emph{Findings of the Association for Computational Linguistics: EMNLP 2021}, pages 591--598.

\bibitem[{Rajpurkar et~al.(2018)Rajpurkar, Jia, and Liang}]{rajpurkar2018know}
Pranav Rajpurkar, Robin Jia, and Percy Liang. 2018.
\newblock Know what you don't know: Unanswerable questions for squad.
\newblock \emph{arXiv preprint arXiv:1806.03822}.

\bibitem[{Rajpurkar et~al.(2016)Rajpurkar, Zhang, Lopyrev, and Liang}]{rajpurkar2016squad}
Pranav Rajpurkar, Jian Zhang, Konstantin Lopyrev, and Percy Liang. 2016.
\newblock Squad: 100,000+ questions for machine comprehension of text.
\newblock \emph{arXiv preprint arXiv:1606.05250}.

\bibitem[{Reiter(1981)}]{reiter1981closed}
Raymond Reiter. 1981.
\newblock On closed world data bases.
\newblock In \emph{Readings in artificial intelligence}, pages 119--140. Elsevier.

\bibitem[{Ruas and Couto(2022)}]{ruas2022nilinker}
Pedro Ruas and Francisco~M Couto. 2022.
\newblock Nilinker: Attention-based approach to nil entity linking.
\newblock \emph{Journal of Biomedical Informatics}, 132:104137.

\bibitem[{Selvaraju et~al.(2017)Selvaraju, Cogswell, Das, Vedantam, Parikh, and Batra}]{selvaraju2017grad}
Ramprasaath~R Selvaraju, Michael Cogswell, Abhishek Das, Ramakrishna Vedantam, Devi Parikh, and Dhruv Batra. 2017.
\newblock Grad-cam: Visual explanations from deep networks via gradient-based localization.
\newblock In \emph{Proceedings of the IEEE international conference on computer vision}, pages 618--626.

\bibitem[{Vaswani et~al.(2017)Vaswani, Shazeer, Parmar, Uszkoreit, Jones, Gomez, Kaiser, and Polosukhin}]{vaswani2017attention}
Ashish Vaswani, Noam Shazeer, Niki Parmar, Jakob Uszkoreit, Llion Jones, Aidan~N Gomez, {\L}ukasz Kaiser, and Illia Polosukhin. 2017.
\newblock Attention is all you need.
\newblock \emph{Advances in neural information processing systems}, 30.

\end{thebibliography}
\bibliographystyle{acl_natbib}
\clearpage
\appendix
\appendix

\section{Differences between Hallucinations in Natural Language Generation and Neural Semantic Parsing}\label{app:nlg-vs-nsp}
Hallucinations manifest differently in Neural Semantic Parsing (NSP) versus Natural Language Generation (NLG) systems. In NSP, hallucinations occur when the predicted logical form or query differs substantively from the gold reference form, despite appearing to be a valid query. This indicates the model fails to accurately capture the full semantic meaning conveyed in the input utterance. However, in NLG, hallucinations arise when the generated text contains false or ungrounded information not directly inferable from the input meaning representation. Whereas NSP hallucinations demonstrate misunderstanding of utterance semantics, NLG hallucinations reflect the model losing contextual grounding to fabricate or hallucinate statements not reasonably justified by reasoning through the implications of the input symbols provided. This suggests brittleness in establishing contextual coherence to match input constraint meanings.

\section{Hallucination Detection Dataset Stats}\label{app:hsf-dataset-stats}
We report the dataset statistics of the Hallucination Simulation Framework in Table \ref{tab:dataset-stats}.
\begin{table}[ht!]
\begin{center}
\begin{tabular}{lc c}
    \toprule
        \textbf{Split} & \textbf{in-ontology} & \textbf{out-of-ontology} \\
    \midrule
        NSP Train & 59,120 & \ \\
        NSP Dev & 19,700  & \ \\ 
        NSP Test & 19,679  & \ \\
    \midrule
        HDD Train & 19,154 & 3,893 \\
        HDD Dev & 546 & 546 \\ 
        HDD Test & 19,679 & 1,467 \\ 
    \bottomrule
\end{tabular}
\end{center}
\caption{Count of sentences for the NSP dataset and for the Hallucination Detection Dataset (HDD) applied to KQA-PRO dataset. We use the term in-ontology and out-of-ontology sentences to refers to the sentences that uses only $\mathcal{O}_{\text{known\_symbols}}$ and $\mathcal{O}_{\text{unknown\_symbols}}$ respectively.}
\label{tab:dataset-stats}
\end{table}

\section{Selection of Unknown symbols}
As mentioned above, we select the symbols for the $\mathcal{O}_{\text{unknown\_symbols}}$ starting from less frequent symbols.
We took all the symbols with at maximum 2 occurrences, this is done b

This is done in order to maintain a good trade off in maximizing the number of 

\section{Out-Of-Domain Dataset Stats}\label{app:ood-stats}
\begin{table}[ht]
\begin{center}
\begin{tabular}{lcc}
    \toprule
    \textbf{Split} & \textbf{NSP Dataset Test} & \textbf{TOP OOD} \\ 
    \midrule
    OOD Test & 17,524  & 35,420 \\ 
    \midrule
\end{tabular}
\end{center}\caption{Size of the TOP v2 out-of-domain dataset used for zero-shot evaluation. The NSP Dataset Test does not include the NSP Errors.}\label{tab:ood-stats}
\end{table}

\section{Out-of-ontology symbols list}
\label{app:out-of-ontology-symbol-list}

Train =['award rationale', 'of', 'separated from', 'quote', 'performer', 'latest date', 'author', 'captain', 'military branch', 'reason for deprecation', 'location', 'has effect', 'doctoral thesis', 'DOI', 'relative to', 'discontinued date', 'applies to part', 'mother', 'quantity', 'conscription number', 'identity of subject in context', 'end cause', 'central bank/issuer', 'dissolved, abolished or demolished', 'employer', 'earliest date', 'located at street address', 'member of political party', 'direction', 'valid in place', 'inventory number', 'series ordinal', 'religious order', 'manufacturer', 'nominee', 'place of marriage', 'creator', 'organizer', 'number of points/goals/set scored', 'nickname', 'number of matches played/races/starts', 'killed by', 'located on street', 'nature of statement', 'position held', 'statement supported by', 'together with', 'street number', 'position played on team / speciality', 'located in or next to body of water', 'instrument', 'doctoral advisor', 'statement disputed by', 'located at street address (DEPRECATED)', 'member of', 'married name', 'stated age at event', 'field of work']

Dev = ['academic degree', 'platform', 'type of kinship', 'present in work', 'appointed by', 'sex or gender', 'image', 'proportion', 'significant event', 'cause of death']

Test = ['catalog code', 'direction relative to location', 'valid in period', 'sourcing circumstances', 'academic major', 'approved by', 'item operated', 'length', 'has cause', 'instance of', 'sRGB color hex triplet', 'operating area', 'conferred by', 'name', 'subject has role', 'applies to jurisdiction', 'prize money', 'conflict', 'head of state', 'affiliation', 'proxy', 'use', 'replaces', 'replaced by', 'writing system', 'located on terrain feature', 'distribution', 'diplomatic mission sent', 'acquisition transaction', 'lyrics by', 'medical condition', 'number of speakers', 'has quality', 'sport number', 'criterion used', 'object has role', 'retrieved', 'basic form of government', 'military rank', 'drafted by', 'timezone offset', 'named as']

\section{Metrics}\label{app:metrics}
We report two metrics to measure the accuracy of our NSP model within the HSF framework.
The MRL Exact Match (EM) consinsts in the ratio between the number of MRL predicted that exactly match with the ground truth MRL over the number of MRLs.
\begin{equation}
    \text{EM} = \frac{1}{|MRLs|}\sum_{k=1}^{|MRLs|} \text{MRL}^{pred}_i == \text{MRL}^{gt}_i
\end{equation}

The Answer Accuracy (AA) instead takes in consideration the retrieved answered from the Knowledge Base and compare it between the ground truth and the predicted one.
\begin{equation}
    \text{AA} = \frac{1}{|MRLs|}\sum_{k=1}^{|MRLs|} \text{ans}_{pred} == \text{ans}_{gt}
\end{equation}
These two metrics differs because sometimes an MRL that does not match with the ground truth can lead to the right answer.
For both metrics, we consider only MRLs that are well-formed and executable, and thus will lead to an answer to be delivered to the customer, as our main concern is preventing the model's users to wrong answers; if an MRL is not executable, it will not lead to answer to be delivered to the user, which in our vision it's better than delivering a wrong (and potentially offensive) answer.

\section{Model Architecture design}\label{hdm-architecture}
\begin{figure}[h!]
\centering
\includegraphics[height=0.49\textheight]{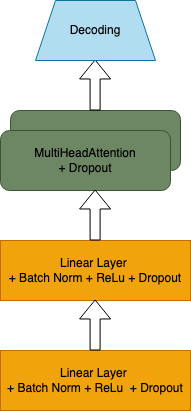}
\caption{Hallucination detection model architecture
}
\label{fig:hdm-architecture}
\end{figure}

In Figure \ref{fig:hdm-architecture}, we report an high level overview of the Hallucination Detection Model architecture, the hyper-parameters used are specified in Section \ref{app:hdm-config}.

\section{Hallucinations in NSP}\label{app:second-hallucination}
\begin{figure}[h!]
\centering
\includegraphics[width=0.49\textwidth]{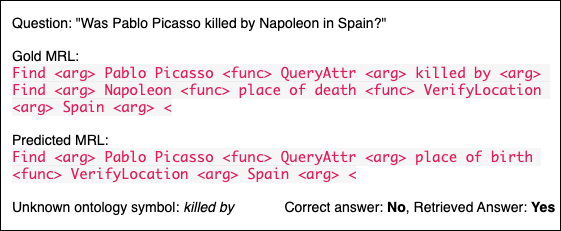}
\caption{We show the output our NSP model trained without a symbol for the concept of ``killed by". Given a 
question that requires this symbol, the model produces a wrong but executable MRL. In that case is it possible to notice that the model avoid to produce the unknown ontology symbol (killed by) and then starts to hallucinate the remaining MRL with wrong known symbols (i.e., place of birth) leading to a complete wrong understanding of the user question. Retrieving a wrong answer. 
}
\label{fig:second-hallucinations}
\end{figure}

In Figure \ref{fig:hallucinated_mrl}, we show how the model hallucinate by omitting portion of the MRL when it encounters the needs of a unknown ontology symbols. However, often, as highlighted in the the discussion, the model replaces the unknown symbols with other known but leading to a complete wrong understanding, thus producing an MRL that is completely hallucinated, we show that behaviour in Figure \ref{fig:second-hallucinations}.
A similar behaviour is observable for NSP Errors. In NSP error the NSP model is under trained on some symbols and then it shows this hallucination behaviour. Instead, in out-of-domain we expect an empty MRL because the model does not have any symbols and syntax to support the out-of-domain user request.

\section{Hallucination Detection Model configuration}\label{app:hdm-config}
We train the HDM using both executable and non-executable MRLs; its training objective is to maximize the number of correctly delivered MRLs and maximize the number of correctly rejected MRLs, regardless of the type of MRLs (e.g., NSP Errors, ontology gap).
The HDM in our Hardware Infrastructure \ref{app:hardware} takes less than a minute to complete each epoch.
In Table \ref{tab:hparams} we report the Hyper-paramters of the best Hallucination Detection Model with Activations + CS.
For sake of brevity, we report the other hyper-parameters configurations in the Github repository.
\begin{table}[ht]
\begin{center}
\begin{tabular}{c|c}
    \toprule
        \textbf{HParams} & \textbf{Value} \\
    \midrule
        Max Epochs & 100 \\
        Optimizer & AdamW \\
        Learning Rate & 1$e^{-3}$  \\ 
        Weight Decay & 1$e^{-3}$ \\ 
        Checkpointing & Max Dev Macro F1-Score \\ 
        Early Stopping & Max Dev Macro F1-Score \\
        Early Stopping Patient & 50 \\ 
        Batch Size & 32 \\
        Non linear activation function & RELU \\
        Loss Function & Cross Entropy \\
        1st layer dim & 1024 \\
        2nd layer dim & 128 \\
        classification head dim & 2 \\
        Precision & fp16 \\
    \bottomrule
\end{tabular}
\end{center}\caption{Hyper-paramters used to train the Hallucination Detection Model.}\label{tab:hparams}
\end{table}

\clearpage
\section{HDD statistics on NSP Errors}\label{app:analysis-nsp-errors}
In Table \ref{tab:analysis-nsp-errors} we report the statistics of the NSP Errors in the Hallucination Detection Dataset.
\begin{table}[ht]
\begin{center}
\begin{tabular}{lc}
    \toprule
    \textbf{Split} & \textbf{NSP Errors percentage} \\ 
    \midrule
    NSP Dataset - Train & 11.01\% \\ 
    NSP Dataset - Dev & 13.66\% \\ 
    NSP Dataset - Test & 10.95\% \\ 
    \bottomrule
\end{tabular}
\end{center}\caption{Percentage of NSP Errors over the executable NSP Dataset. Computed using the NSP model at inference time by comparing the predictions with the ground truth.}\label{tab:analysis-nsp-errors}
\end{table}

\section{HDD statistics on Executable MRLs}\label{app:executableMRL-stats}
In Table \ref{tab:executableMRL-stats} we report the percentage of executable MRLs in the Hallucination Detection Dataset w.r.t the KQA-PRO BART inference trained on the NSP in-ontology dataset.
\begin{table}
\begin{center}
\begin{tabular}{lc}
    \toprule
    \textbf{Split} & \textbf{Executable} \\ 
    \midrule
    \textbf{HDD Train} & \\
    \midrule
    in-ontology & 92.69\% \\ 
    out-of-ontology & 42.97\% \\ 
    \midrule
    \textbf{HDD Dev}  & \\
    \midrule
    in-ontology & 92.49\%  \\ 
    out-of-ontology & 30.04\%  \\
    \midrule
    \textbf{HDD Test}  & \\
    \midrule
    in-ontology & 92.63\%  \\ 
    out-of-ontology & 46.27\% \\
    \bottomrule
\end{tabular}
\end{center}\caption{Percentage of executable MRLs in HDD, after KQA-PRO BART inference. We use the term in-ontology and out-of-ontology sentences to refers to the sentences that uses only $\mathcal{O}_{\text{known\_symbols}}$ and $\mathcal{O}_{\text{unknown\_symbols}}$ respectively.}\label{tab:executableMRL-stats}
\end{table}

\section{Hardware Infrastructure}\label{app:hardware}
We performed all the experiments on a x86-64 architecture with 748GB of RAM, 4x 24-core CPU Intel Xeon Platinum 8175M, and a single NVIDIA V100 with 32GB of VRAM.

\section{KQA-PRO Bart hyper-parameters}\label{app:kqa-pro-hparams}
To fine-tune the BART model on the KQA-PRO dataset, we stick with the same hyper-parameters used by the \citet{cao2022kqa}.
Below are the only changes in hyper-parameters we have made.
We reduce the number of epochs from 25 to 3, which we found to be sufficient to achieve high performance while vastly reducing the training time. 
We also enable beam search with a beam size of 4, to compute the aforementioned Confidence Score feature.

\section{Precision and Recall}\label{app:precision-recall}
We report the Macro Precision and Macro Recall performance in Tables \ref{tab:out-of-ontology-precision-recall} \ref{tab:nsp-errors-precision-recall}, \ref{tab:out-of-domain-precision-recall} for out-of-ontology, NSP Errors, and OOD.
\begin{table*}[t]
    \centering
    \begin{tabular}[t]{lcc}
        \toprule
        \textsc{\textbf{Exp name - out-of-ontology}}
        & \textbf{Precision}
        & \textbf{Recall} \\
        \midrule
            No-Filter (Baseline) & 0.480 & 0.5 \\
            Threshold CS 98.5\% (Baseline) & 0.479 & 0.482 \\
            Threshold MCD (Baseline) & 0.477 & 0.430 \\
            Activations$^{HDM}$ & 0.557 ± 0.110 & 0.504 ± 0.007 \\
            CS$^{HDM}$  & 0.671 ± 0.056 & 0.663 ± 0.034 \\
            MCD$^{HDM}$ & 0.591 ± 0.208 & 0.500 ± 0.001 \\
            CS + MCD$^{HDM}$ & 0.654 ± 0.027 & 0.657 ± 0.019 \\
            Activations + CS$^{HDM}$  & \textbf{0.682 ± 0.041} & \textbf{0.717 ± 0.019} \\
            Activations + MCD$^{HDM}$ & 0.593 ± 0.159 & 0.502 ± 0.007 \\
            Activations + CS + MCD$^{HDM}$ & 0.642 ± 0.033 & 0.691 ± 0.018 \\
        \bottomrule
    \end{tabular}
    \caption{We report the Macro F1-Score ($\uparrow$ is better) in out-of-ontology detection. We have repeated the train of the HDM using 10 random seeds, we report the mean of the scores along with their standard deviation. These features are combined (+) concatenating their vector representations.}
    \label{tab:out-of-ontology-precision-recall}
\end{table*}

\begin{table*}[t]
    \centering
    \begin{tabular}[t]{lcc}
        \toprule
        \textsc{\textbf{Exp name - NSP Errors}}
        & \textbf{Precision}
        & \textbf{Recall} \\
        \midrule
            No-Filter (Baseline) & 0.444 & 0.500 \\
            Threshold CS 98.5\% (Baseline) & 0.706 & 0.627 \\
            Threshold MCD (Baseline) & 0.442 & 0.436 \\
            Activations$^{HDM}$ & 0.547 ± 0.060 & 0.501 ± 0.002 \\
            CS$^{HDM}$  & 0.698 ± 0.009 & 0.571 ± 0.018 \\
            MCD$^{HDM}$ & 0.444 ± 0.002 & 0.500 ± 0.004 \\
            CS + MCD$^{HDM}$ & 0.695 ± 0.006 & 0.594 ± 0.016 \\
            Activations + CS$^{HDM}$  & \textbf{0.712 ± 0.007} & 0.619 ± 0.029 \\
            Activations + MCD$^{HDM}$ & 0.515 ± 0.044 & 0.501 ± 0.001 \\
            Activations + CS+ MCD$^{HDM}$ & 0.705 ± 0.008 & \textbf{0.641 ± 0.034} \\
        \bottomrule
    \end{tabular}
    \caption{We report the Macro Precision and Recall ($\uparrow$ is better) in NSP Errors detection, NSP Error detection. We have repeated the train of the HDM using 10 random seeds, we report the mean of the scores along with their standard deviation. These features are combined (+) concatenating their vector representations.}
    \label{tab:nsp-errors-precision-recall}
\end{table*}

\begin{table*}[t]
    \centering
    \begin{tabular}[t]{lcc}
        \toprule
        \textsc{\textbf{Exp name - out-of-domain}}
        & \textbf{Precision}
        & \textbf{Recall} \\
        \midrule
            No-Filter (Baseline) & 0.436 & 0.500 \\
            Threshold CS 98.5\% (Baseline) & 0.434 & 0.482 \\
            Threshold MCD (Baseline) & 0.427 & 0.429 \\
            Activations$^{HDM}$ & 0.479 ± 0.116 & 0.499 ± 0.002 \\
            CS$^{HDM}$  & 0.614 ± 0.039 & 0.632 ± 0.052 \\
            MCD$^{HDM}$ & 0.644 ± 0.019 & 0.563 ± 0.151 \\
            CS + MCD$^{HDM}$ & 0.595 ± 0.061 & 0.534 ± 0.021 \\
            Activations + CS$^{HDM}$  & \textbf{0.760 ± 0.108} & \textbf{0.662 ± 0.071} \\
            Activations + MCD$^{HDM}$ & 0.447 ± 0.018 & 0.498 ± 0.002 \\
            Activations + CS+ MCD$^{HDM}$ & 0.671 ± 0.101 & 0.599 ± 0.063 \\
        \bottomrule
    \end{tabular}
    \caption{We report the Macro Precision and Recall ($\uparrow$ is better) in zero-shot out-of-domain detection. We have repeated the train of the HDM using 10 random seeds, we report the mean of the scores along with their standard deviation. These features are combined (+) concatenating their vector representations.}
    \label{tab:out-of-domain-precision-recall}
\end{table*}

\end{document}